\documentclass[10pt,twocolumn,letterpaper]{article}

\usepackage[pagenumbers]{iccv} % To force page numbers, e.g. for an arXiv version

\usepackage{{xspace}}
\newcommand{\ct}[1]{\multicolumn{1}{c}{#1}}

\newcommand{\cls}{[\texttt{CLS}]\xspace}

\definecolor{iccvblue}{rgb}{0.21,0.49,0.74}
\usepackage[pagebackref,breaklinks,colorlinks,allcolors=iccvblue]{hyperref}
\usepackage{multicol, multirow}
\usepackage{colortbl}

\title{Rethinking the Use of Vision Transformers for AI-Generated Image Detection}

\author{
NaHyeon Park\\
KAIST AI\\
\small{julia19@kaist.ac.kr} 
\and
Kunhee Kim\\
KAIST AI\\
\small{kunhee.kim@kaist.ac.kr}
\and
Junsuk Choe\\
Sogang University\\
\small{jschoe@sogang.ac.kr}
\and
Hyunjung Shim\\
KAIST AI\\
\small{kateshim@kaist.ac.kr}
}

\begin{document}
\maketitle
\begin{abstract}
Rich feature representations derived from CLIP-ViT have been widely utilized in AI-generated image detection. While most existing methods primarily leverage features from the final layer, we systematically analyze the contributions of layer-wise features to this task. Our study reveals that earlier layers provide more localized and generalizable features, often surpassing the performance of final-layer features in detection tasks. Moreover, we find that different layers capture distinct aspects of the data, each contributing uniquely to AI-generated image detection. Motivated by these findings, we introduce a novel adaptive method, termed \textbf{\underline{M}}ixture \textbf{\underline{o}}f \textbf{\underline{L}}ayers for AI-generated image \textbf{\underline{D}}etection (MoLD), which dynamically integrates features from multiple ViT layers using a gating-based mechanism. Extensive experiments on both GAN- and diffusion-generated images demonstrate that MoLD significantly improves detection performance, enhances generalization across diverse generative models, and exhibits robustness in real-world scenarios. Finally, we illustrate the scalability and versatility of our approach by successfully applying it to other pre-trained ViTs, such as DINOv2. Code is available in \url{https://github.com/nahyeonkaty/mold}.
\end{abstract}
    
\section{Introduction}
\label{sec:intro}

The rapid advancement of generative models, such as generative adversarial networks (GANs)~\cite{goodfellow2020generative} and diffusion models~\cite{sohl2015deep, ho2020denoising}, has significantly improved the realism and fidelity of synthesized images. 
In particular, the emergence of text-to-image diffusion models~\cite{nichol2021glide, ramesh2021zero, rombach2022high, podell2023sdxl} has revolutionized image synthesis, enabling the generation of highly detailed visuals from natural language descriptions. 
However, as AI-generated images become highly sophisticated, distinguishing them from real images has become increasingly challenging, raising critical concerns regarding privacy~\cite{liu2020global, frank2020leveraging, aneja2020generalized}, digital authenticity~\cite{guan2022delving, yan2023ucf, liu2024forgery}, and the proliferation of misinformation~\cite{nguyen2024laa, lin2024preserving}. 
Consequently, the development of reliable AI-generated image detection techniques has become a crucial area of research~\cite{wang2020cnn, tan2023learning, tan2024rethinking, epstein2023online, lanzino2024faster}.

A key challenge in AI-generated image detection is ensuring generalizability across different generative models. 
An effective detection system should not only recognize images produced by known models but also generalize to images from unseen generative models. 
Recent work by \citet{ojha2023towards} has demonstrated that the pre-trained CLIP~\cite{radford2021learning}, a vision-language model trained on a large-scale dataset of 400 million real image-text pairs, can be effectively utilized for AI-generated image detection. Their findings indicate that a simple classification head built upon CLIP-derived features exhibits strong generalization capabilities across various generative models.
Building on this foundation, subsequent studies~\cite{liu2024mixture, chen2024drct, cozzolino2024raising} have further investigated CLIP-based detectors. 
However, these approaches predominantly leverage features from the final layers of CLIP’s vision transformer (ViT), which may constrain their effectiveness in capturing richer representational patterns. 

In this paper, we address a critical research question: Is relying exclusively on the last-layer features of CLIP’s ViT the optimal approach for effective AI-generated image detection? 
This question is motivated by prior studies on ViTs~\cite{raghu2021vision, park2023self}, which suggest that earlier layers capture both local and global information, whereas later layers predominantly encode high-level semantic representations. 
Given that AI-generated image detection is more closely associated with local image attributes rather than high-level semantic understanding~\cite{ojha2023towards, tan2023learning, tan2024rethinking, wang2023dire, sinitsa2024deep, chai2020makes}, we hypothesize that relying solely on last-layer features, which primarily capture high-level semantics, may not be optimal for this task. 
Instead, we investigate the use of earlier-layer features, positing that different layers may encode more relevant information for distinguishing AI-generated images.

To investigate this hypothesis, we conduct a comprehensive layer-wise analysis of CLIP-ViT by training classifiers on features extracted from different layers while keeping the CLIP backbone frozen. 
Specifically, each classifier is trained using features from a specific layer index and evaluated for AI-generated image detection. 
We then assess the performance of these classifiers on images generated by diverse generative models. 
Our analysis yields several key observations:
\begin{itemize}
\item (O1) Features from very early layers exhibit suboptimal performance due to their highly localized nature. In contrast, mid-layer features achieve superior detection accuracy compared to features from the final layer.
\item (O2) The optimal layer-wise feature for AI-generated image detection varies across datasets, suggesting that no single layer universally provides the best discriminative features.
\end{itemize}

To further investigate the degree of information overlap among different layers, we analyze the proportion of commonly misclassified samples to assess the consistency of classifiers trained on features extracted from different layers.  
\begin{itemize}
\item (O3) Our findings demonstrate that different layers encode distinct and complementary information relevant to AI-generated image detection. This suggests that the features learned at each layer contribute uniquely to the classification process rather than being redundant.
\end{itemize}

Based on these insights, we propose \textbf{\underline{M}}ixture \textbf{\underline{o}}f \textbf{\underline{L}}ayers for AI-generated image \textbf{\underline{D}}etection (MoLD), a novel adaptive approach inspired by the Mixture-of-Experts paradigm. 
Specifically, MoLD aggregates \cls tokens from multiple layers using a data-dependent gating network that dynamically assigns weights based on input characteristics. Each layer's features are projected into a shared representational space and combined adaptively, allowing MoLD to leverage the distinct and complementary information encoded across different ViT layers effectively.

Extensive experiments, evaluated on GAN-based images from ForenSynths~\cite{wang2020cnn} and diffusion-generated images from the GenImage~\cite{zhu2023genimage} dataset, demonstrate that our proposed method significantly enhances detection performance with high generalization capability across diverse generative models. 
Furthermore, we assess the robustness of our method under various scenarios, including adversarial perturbations and different training dataset sizes. 
Finally, we show that our MoLD consistently improves detection performance across different pre-trained ViTs, thereby broadening the applicability of pre-trained feature representations for AI-generated image detection.

In summary, our main contributions are as follows: 
\begin{itemize} 
    \item We conduct a detailed layer-wise analysis of CLIP-based ViTs for AI-generated image detection, demonstrating that features from different layers can be effective for this task. 
    \item We show that different ViT layers contribute distinct information to AI-generated image detection, enhancing the robustness of detection systems with diverse feature representations. 
    \item We propose MoLD, an adaptive, gating-based method that aggregates features from multiple ViT layers, significantly improving detection performance.
    \item We validate our approach through extensive experiments on multiple datasets, demonstrating its effectiveness and generalizability across diverse generative models and pre-trained ViTs. 
\end{itemize}

\section{Related Work}
\label{sec:rwork}

\subsection{AI-generated image detection}
The field of fake image detection has been extensively studied~\cite{wang2020cnn, chai2020makes, sinitsa2024deep, zhong2023patchcraft, grommelt2024fake}, driven by the rapid advancements in generative models. 
Many studies~\cite{wang2023dire,  liu2024forgery, cazenavette2024fakeinversion, luo2024lare, ricker2024aeroblade, wang2024trace, ma2023exposing} have investigated the distinguishing characteristics between real and synthetic images, and utilized them to train fake image detector.
In contrast to these methods, \citet{ojha2023towards} argued that detectors focusing solely on fake-specific features tend to introduce bias, leading to reduced generalizability. 
They proposed that a more effective approach for universal fake image detection involves utilizing feature representations not exclusively trained for this task. Their findings suggest that the feature space of a pre-trained CLIP-ViT model is particularly well-suited for generalizable fake image detection. 
This discovery has spurred further research~\cite{cozzolino2024raising, liu2024mixture, chen2024drct} into the application of CLIP-ViT for fake image detection. 
For example, \citet{chen2024drct} extracted features from CLIP and adopted contrastive learning framework that utilizes images reconstructed by diffusion models as negative samples to enhance fake image detection. 
Unlike previous studies that primarily focus on the final layers of neural networks, our work investigates the effectiveness of different layers for AI-generated image detection.

\subsection{Layer-wise characteristics of ViTs}
Numerous studies~\cite{naseer2021intriguing, heo2021rethinking, park2022vision, park2023self} have examined the layer-wise characteristics of Vision Transformers (ViTs)~\cite{dosovitskiy2020image}.
\citet{raghu2021vision} demonstrated that, given a sufficiently large dataset, the lower (i.e., earlier) layers of ViTs attend to both local and global regions, whereas the higher (i.e., later) layers primarily capture global context.
In the context of transfer learning, \citet{yosinski2014transferable} showed that early layers learn more generalizable features applicable across multiple tasks, while higher layers encode representations that are more specialized to the pre-training objective.
Leveraging these distinct layer-wise characteristics, a number of studies~\cite{alain2016understanding, jiang2018predicting, evci2022head2toe, lee2022surgical, heo2024llms, huang2019enhancing, adler2020cross} have explored the role of ViT layers in various applications.
For instance, \citet{uselisintermediate} demonstrated that the intermediate layers of a ViT play a crucial role in out-of-distribution (OOD) generalization, underscoring the significance of representations learned at different depths.
Building upon these prior investigations, our study examines the role of features extracted from different layers of a pre-trained ViT.
Specifically, we evaluate their effectiveness in AI-generated image detection, an area that has not been explicitly explored in prior work.
\section{Approach}
\label{sec:approach}
%------------------------------------------
\begin{figure}[t]
    \centering
    \includegraphics[width=0.9\linewidth]{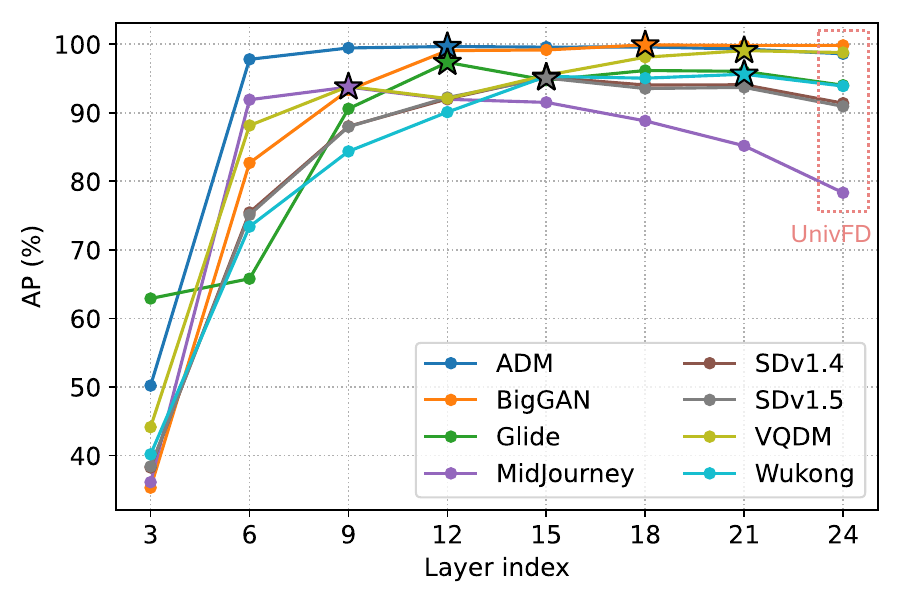}
      \caption{\textbf{Layer-wise performance in AI-generated image detection.} We train individual classifiers using pre-trained CLIP:ViT-L/14 features from different layers of a model on the GenImage-ADM training set. The performance of each layer-specific classifier is then evaluated on various test subsets, and the average precision is reported. The results indicate that the optimal performance (marked as star for each test subset) varies across datasets, with mid-layer features generally outperforming those from earlier or later layers.}
    \label{fig:motivation1}
\end{figure}
%------------------------------------------

\subsection{Motivation}
Our work is inspired by previous studies on Vision Transformers (ViTs)~\cite{raghu2021vision, park2023self}, which suggest that earlier layers capture both local and global information, while later layers primarily encode high-level semantic representations. 
Since AI-generated image detection is a task that has low relevance with high-level semantics of an image~\cite{tan2023learning, tan2024rethinking, wang2023dire, sinitsa2024deep, chai2020makes}, relying exclusively on the features from the final layer may lead to suboptimal utilization of the feature space provided by CLIP-ViT.
Indeed, we find that the final layers are highly entangled with semantic information (see Supp. A.1).
In light of this, we systematically examine the characteristics and effectiveness of features from different layers in the context of AI-generated image detection.

\subsection{Prior Analysis}
\label{sec:analysis}

\textbf{Experiment design.}
To investigate this in detail, a systematic layer-wise experiment is designed to examine  layer-wise contribution of each group of transformer block's features for AI-generated image detection.
Specifically, we trained classifiers using features extracted from individual layers of the CLIP:ViT/L-14 model, which consists of 24 transformer layers. 
Given that the layers in the ViTs are sequentially structured, we grouped them into sets of three consecutive layers to simplify the analysis and improve computational efficiency (All layers of CLIP:ViT/B-16 in Supp. A.2).
It is important to note that the ViT backbone was kept frozen throughout this process, making only the classification head learnable.
For each layer index $i$, we extracted the feature representations from the output of the $i$-th transformer block and passed these features through a classifier to obtain the corresponding logits. 
Specifically, we trained a total of eight classifiers, corresponding to the layers with indices $i \in \{3, 6, 9, 12, 15, 18, 21, 24\}$. 
The classifiers were trained on the training subset, and each trained detector was evaluated on the test subset of GenImage~\cite{zhu2023genimage}.

\noindent\textbf{Using mid-layer features leads to performance gain.}
We begin by evaluating the performance of trained layer-wise classifiers for AI-generated image detection. The results of this experiment are presented in Figure~\ref{fig:motivation1}. 
Our findings indicate that features extracted from the earliest layers exhibit suboptimal performance, likely due to their highly localized nature. 
Notably, we observe that features from mid-level layers outperform those from the final layer, which is the approach employed in UnivFD~\cite{ojha2023towards}.
We also have extended this analysis to other settings: different backbone and datasets. Please refer to Supp. A.2 for additional settings.
The results confirm that this trend generally holds across different datasets and generative model. 

%------------------------------------------
\begin{figure}[t]
    \centering
    \includegraphics[width=\linewidth]{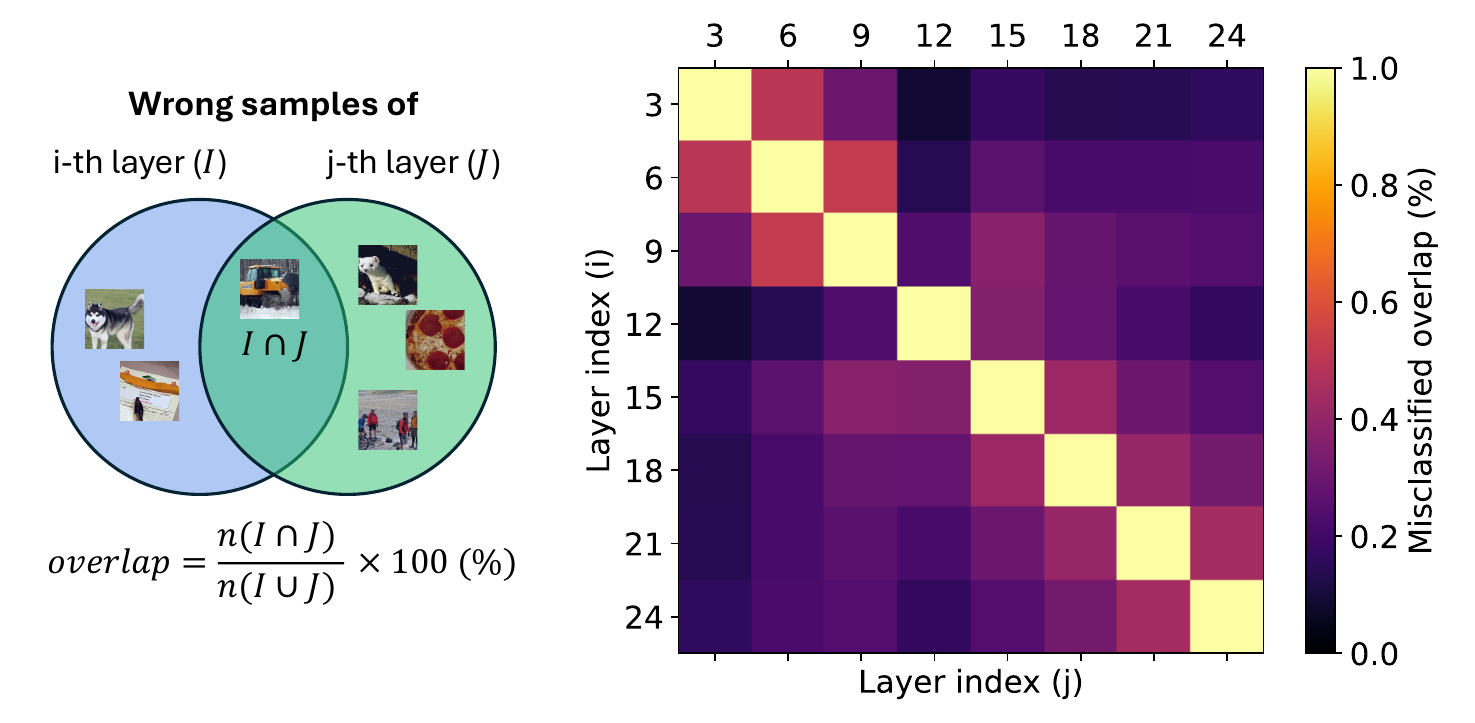}
    \caption{\textbf{Overlap in misclassified samples across layers.} To analyze the decision-making process of classifiers trained on features from different layers, we measure the proportion of commonly misclassified samples. The results indicate that the overlap in misclassified samples across layers is relatively low, suggesting that each layer captures distinct aspects of the input data. This highlights the complementary nature of multi-layer representations and underscores the importance of integrating features from all layers for robust fake image detection.}
    \label{fig:motivation2}
\end{figure}
%------------------------------------------

\noindent\textbf{Optimal layer features vary across datasets.}
As illustrated in Figure~\ref{fig:motivation1}, the highest performance of the trained fake image detector was achieved by utilizing features extracted from layers other than the final layer.
However, the optimal layer for feature extraction varies across different test datasets. For instance, on the ADM test set, the 12-th layer provides the best performance (indicated by the blue star), while on the BigGAN test set, the 18-th layer performs the best (indicated by the orange star). 
This variation raises an important question: which specific network layer yields the most informative features for optimal detection performance? 
Since the optimal layer differs across datasets, a fixed-layer approach may not be universally effective.
% Therefore, an universal approach is necessary to ensure consistently high detection accuracy across diverse datasets.

%------------------------------------------
\begin{figure}[t]
    \centering
    \includegraphics[width=0.975\linewidth]{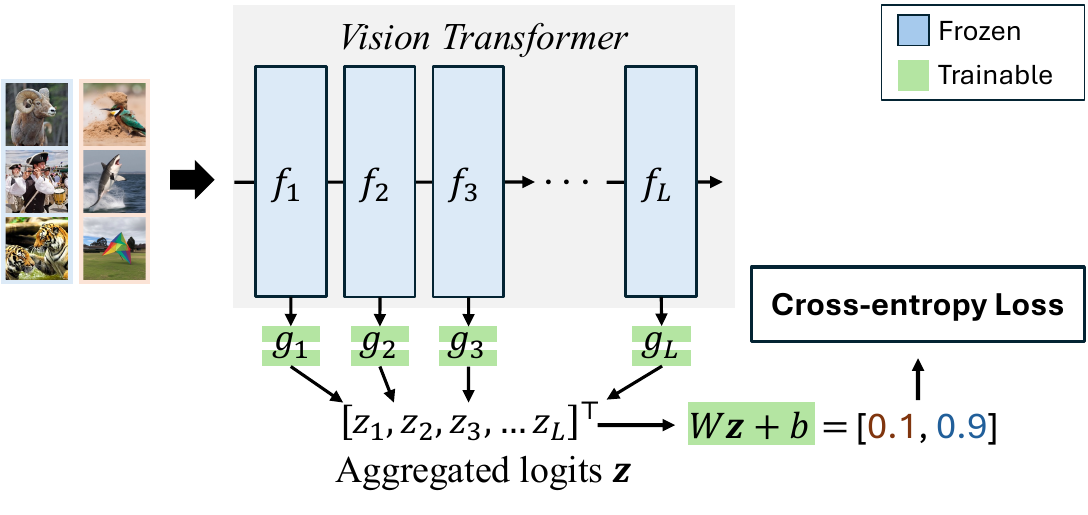}
    \caption{\textbf{Overview of our \texttt{MoLD}.} Our approach fully leverages the features of a pre-trained Vision Transformer by aggregating the \cls token embeddings from all transformer layers. A dedicated lightweight network is applied at each layer to generate layer-wise predictions. These predictions are then processed by a learnable classification head to produce the final prediction. The layer-wise networks $g_i$ and the classification head are jointly trained using binary cross-entropy loss. Note that the ViT backbone remains frozen throughout the entire training process.
    }
    \label{fig:method}
\end{figure}
%------------------------------------------

\noindent\textbf{Features of each ViT layer has different contributions.}
To gain a deeper understanding of the decision-making process at various layers of the model, we conducted an experiment to assess whether classifiers trained on features from different layers exhibit similar or divergent classification errors. 
Specifically, we measured the proportion of misclassified samples for each layer-specific classifier. 
The results, presented in Figure~\ref{fig:motivation2}, demonstrate that the overlap in misclassified samples across layers is low (additional results in Supp. A.3). 
This finding suggests that classifiers trained on features from different layers tend to make distinct errors, indicating that each layer captures unique aspects of the data. Consequently, this highlights the value of leveraging multi-layer features when constructing a robust fake image detector. 
By integrating features from multiple layers, we can fully utilize the pre-trained CLIP-ViT's representations, thereby enhancing the detector's capacity to capture a broader range of discriminative information.
%% 이 관찰이 dataset / 모델이 바뀌어도 경향성이 유지된다는 것을 명시하고, 
%% 이것에 대한 추가적인 실험을 supple에 제시.

%--------------------TAB: PROGAN --------------------
\begin{table*}[ht]
 % \vspace{-0.2 cm}
    \centering
\resizebox{\textwidth}{!}{
    \begin{tabular}{l  r r r r r r r r r r r r r r r r | r r}
    \toprule
       \multirow{2}*{Method} & \multicolumn{2}{c}{ProGAN}& \multicolumn{2}{c}{StyleGAN}& \multicolumn{2}{c}{StyleGAN2}& \multicolumn{2}{c}{BigGAN}& \multicolumn{2}{c}{CycleGAN}& \multicolumn{2}{c}{StarGAN}& \multicolumn{2}{c}{GauGAN}& \multicolumn{2}{c|}{Deepfake}& \multicolumn{2}{c}{Mean}\\

         \cmidrule{2-19} ~   & \ct{ACC} & \ct{AP} & \ct{ACC} & \ct{AP} & \ct{ACC} & \ct{AP} & \ct{ACC} & \ct{AP} & \ct{ACC} & \ct{AP} & \ct{ACC} & \ct{AP} & \ct{ACC} & \ct{AP}  & \ct{ACC} & \ct{AP} & \multicolumn{1}{|c}{ACC} & \ct{AP} \\ \midrule
        CNNDetection~\cite{wang2020cnn} & 91.4 & 99.4 & 63.8 & 91.4 & 76.4 & 97.5 & 52.9 & 73.3 & 72.7 & 88.6 & 63.8 & 90.8 & 63.9 & 92.2 & 51.7 & 62.3 & 67.1  & 86.9 \\  
        PatchFor~\cite{chai2020makes} & 97.8 & 100.0 & 82.6 & 93.1 & 83.6 & 98.5 & 64.7 & 69.5 & 74.5 & 87.2 & 100.0 & 100.0 & 57.2 & 55.4 & 85.0 & 93.2 & 80.7 & 87.1  \\
        LGrad~\cite{tan2023learning} & 99.9 & 100.0& 94.8 & 99.9 & 96.0 & 99.9 & 82.9 & 90.7 & 85.3 & 94.0 & 99.6 & 100.0& 72.4 & 79.3 & 58.0 & 67.9 & 86.1 & 91.5 \\
        UnivFD~\cite{ojha2023towards} & 99.7 & 100.0 & 89.0 & 98.7 & 83.9 & 98.4 & 90.5 & 99.1 & 87.9 & 99.8 & 91.4 & 100.0& 89.9 & 100.0 & 80.2 & 90.2 & 89.1 & 98.3 \\
        DIRE~\cite{wang2023dire} & 98.3 & 99.9 & 72.5 & 94.3 & 66.3 & 97.7 & 59.1 & 79.2 & 66.2 & 78.2 & 92.8 & 100.0 & 54.9 & 72.2 & 87.0 & 97.1 & 74.6 & 89.8 \\
        NPR~\cite{tan2024rethinking} & 99.8 & 100.0& 96.3 & 99.8 & 97.3 & 100.0& 87.5 & 94.5 & 95.0 & 99.5 & 99.7 & 100.0 & 86.6 & 88.8 & 77.4 & 86.2 & \textbf{92.5} & 96.1 \\
        DRCT (Conv-B)~\cite{chen2024drct} & 98.6 & 99.9 & 76.2 & 96.2 & 60.8 & 96.2 & 86.0 & 75.7 & 96.9 & 98.0 & 61.4 & 94.1 & 82.6 & 99.5 & 33.9 & 75.3 & 74.6 & 91.9 \\
        DRCT (CLIP-L)~\cite{chen2024drct} & 98.4 & 99.9 & 82.8 & 95.0 & 74.5 & 95.4 & 87.9 & 96.7 & 92.2 & 98.0 & 83.2 & 95.8 & 98.5 & 99.9 & 42.4 & 78.6 & 82.5 & 94.9 \\
        \midrule
        % Ours (simple) & 100.0 & 100.0 & 98.7 & 100.0 & 89.9 & 99.3 & 99.1 & 100.0 & 98.5 & 100.0 & 98.3 & 100.0 & 98.7 & 99.9 & 71.5 & 97.5 & \testbf{94.3} & \textbf{99.6} \\
        \rowcolor{gray!25} MoLD (Ours) & 99.9 & 100.0 & 91.1 & 99.8 & 86.0 & 99.8 & 98.4 & 100.0 & 98.1 & 99.9 & 99.1 & 100.0 & 99.7 & 100.0 & 58.7 & 96.5 & 91.4 & \textbf{99.5} \\
        
\bottomrule
    \end{tabular}
}
  \caption{\textbf{Evaluation on the ForenSynths~\cite{wang2020cnn}.} We follow the same experimental setting of \citet{tan2024rethinking}, and the values in this table are borrowed from their paper. To incorporate more recent works, we train DIRE and DRCT and add their results to the table. We present accuracy (ACC) and average precision (AP) in percentage for each test set and compute the overall average.}
  \label{tab:progan}
     % \vspace{-0.25 cm}
\end{table*}
%-------------------- END OF TAB --------------------

%--------------------TAB: GENIMAGE-ADM --------------------
\begin{table*}[ht]
 % \vspace{-0.2 cm}
    \centering
\resizebox{\textwidth}{!}{
    \begin{tabular}{l  r r r r r r r r r r r r r r r r | r r}
    \toprule
       \multirow{2}*{Method} & \multicolumn{2}{c}{ADM}& \multicolumn{2}{c}{BigGAN}& \multicolumn{2}{c}{GLIDE}& \multicolumn{2}{c}{Midjourney}& \multicolumn{2}{c}{SDv1.4}& \multicolumn{2}{c}{SDv1.5}& \multicolumn{2}{c}{VQDM}& \multicolumn{2}{c|}{Wukong}& \multicolumn{2}{c}{Mean}\\

         \cline{2-19} ~   & \ct{ACC} & \ct{AP} & \ct{ACC} & \ct{AP} & \ct{ACC} & \ct{AP} & \ct{ACC} & \ct{AP} & \ct{ACC} & \ct{AP} & \ct{ACC} & \ct{AP} & \ct{ACC} & \ct{AP}  & \ct{ACC} & \ct{AP}  & \multicolumn{1}{|c}{ACC} & \ct{AP} \\ \midrule
        CNNDetection~\cite{wang2020cnn} & 99.7 & 100.0 & 87.3 & 99.3 & 96.2 & 99.6 & 64.6 & 90.1 & 57.8 & 86.2 & 58.0 & 87.1 & 80.0 & 97.7 & 54.8 & 79.1 & 74.8 & 92.4 \\ 
        PatchFor~\cite{chai2020makes} & 100.0 & 100.0 & 50.0 & 97.4 & 98.1 & 100.0 & 51.3 & 87.7 & 50.0 & 69.2 & 50.0 & 68.7 & 100.0 & 100.0 & 50.0 & 71.1 & 68.7 & 86.8 \\
        LGrad~\cite{tan2023learning} & 99.9 & 100.0 & 57.2 & 98.0 & 94.2 & 99.6 & 62.1 & 93.5 & 56.5 & 89.4 & 56.4 & 89.7 & 92.4 & 99.8 & 53.9 & 81.9 & 71.6 & 94.0 \\
        UnivFD~\cite{ojha2023towards} & 90.6 & 97.1 & 91.7 & 99.1 & 79.0 & 88.2 & 61.8 & 69.5 & 80.3 & 89.2 & 79.5 & 88.4 & 90.8 & 98.5 & 84.8 & 93.3 & 82.3 & 90.4 \\
        DIRE~\cite{wang2023dire} & 99.5 & 100.0 & 67.6 & 94.8 & 94.3 & 99.1 & 62.7 & 88.1 & 56.3 & 80.2 & 56.3 & 80.7 & 73.4 & 95.7 & 54.2 & 73.4 & 70.5 & 89.0 \\
        NPR~\cite{tan2024rethinking} & 100.0 & 100.0 & 51.2 & 98.6 & 96.1 & 99.8 & 64.4 & 90.9 & 52.9 & 77.9 & 52.7 & 78.6 & 74.9 & 97.8 & 52.3 & 73.8 & 63.5 & 88.2 \\
        DRCT (Conv-B)~\cite{chen2024drct} & 99.6 & 100.0 & 65.9 & 96.3 & 96.5 & 99.8 & 67.0 & 94.9 & 68.9 & 96.4 & 68.2 & 96.5 & 81.5 & 98.8 & 63.9 & 93.9 & 76.4 & 97.1 \\
        DRCT (CLIP-L)~\cite{chen2024drct} & 95.4 & 99.3 & 86.5 & 97.3 & 93.5 & 99.0 & 57.6 & 77.4 & 71.0 & 91.4 & 70.1 & 90.4 & 91.3 & 98.5 & 69.5 & 89.8 & 79.4 & 92.9 \\
        \midrule
        % Ours (simple)    & 96.76 & 99.66 & 96.21 & 99.06 & 92.35 & 97.34 & 81.55 & 91.96 & 80.87 & 92.01 & 80.62 & 92.22 & 80.96 & 92.07 & 76.76 & 90.09 & 85.76 & 94.30 \\
        \rowcolor{gray!25} MoLD (Ours) & 99.3 & 100.0 & 83.3 & 97.9 & 91.1 & 99.0 & 76.3 & 95.2 & 88.2 & 98.5 & 87.0 & 98.3 & 93.5 & 99.2 & 86.5 & 98.1 & \multicolumn{1}{|r}{\textbf{88.2}} & \textbf{98.2} \\
    \bottomrule
    \end{tabular}
}
  \caption{\textbf{Evaluation on the GenImage~\cite{zhu2023genimage}.} To assess the effectiveness of our approach on recent generative models, we evaluate both baseline methods and our proposed approach using the GenImage dataset, which includes samples from GAN-based and Diffusion-based models. To demonstrate generalizability, we train the fake image detectors on the training set of GenImage-ADM (the first model in alphabetical order) and evaluate their performance on the test sets of different generative models. We report accuracy (ACC) and average precision (AP) in percentage (\%) for each test set and compute the overall average.}
  \label{tab:genimage-adm}
     % \vspace{-0.25 cm}
\end{table*}
%-------------------- END OF TAB --------------------
%--------------------TAB: CROSSVAL (GENIMAGE) --------------------
\begin{table}[!ht]
 % \vspace{-0.2 cm}
    \centering
\resizebox{\linewidth}{!}{
    \begin{tabular}{l r r r r r r r r r r r r r r r r | r r}
    \toprule
       & \ct{ADM} & \ct{BigGAN} & \ct{GLIDE} & \ct{Midjourney} & \ct{SDv1.4} & \ct{SDv1.5} & \ct{VQDM} & \ct{Wukong} \\
        \midrule
        ADM & 100.0 & 97.9 & 99.0 & 95.2 & 98.5 & 98.3 & 99.2 & 98.1 \\ 
        BigGAN & 78.4 & 99.9 & 95.5 & 76.1 & 94.0 & 94.5 & 97.5 & 90.7 \\ 
        GLIDE & 94.0 & 95.7 & 100.0 & 90.4 & 97.6 & 97.4 & 95.3 & 95.9 \\ 
        Midjourney & 71.4 & 87.6 & 86.6 & 98.7 & 86.9 & 87.3 & 51.9 & 77.8 \\ 
        SDv1.4 & 80.5 & 81.6 & 92.4 & 95.8 & 99.9 & 99.9 & 89.0 & 98.6 \\ 
        SDv1.5 & 79.1 & 75.8 & 92.9 & 95.5 & 99.9 & 99.9 & 90.1 & 98.9 \\ 
        VQDM & 92.7 & 97.8 & 97.9 & 97.4 & 99.8 & 99.6 & 100.0 & 99.8 \\ 
        Wukong & 87.0 & 71.9 & 93.2 & 89.0 & 99.9 & 99.9 & 94.4 & 99.7 \\ 
\bottomrule
    \end{tabular}
}
  \caption{\textbf{Cross-validation on different train sets.} To assess the generalizability of our approach across different training images generated by various source generative models, we train our model on different subsets of the training data. The left column indicates the training set derived from GenImage. We then evaluate the model on all other test subsets and report the average precision (AP) for each combination of training and test sets. The results demonstrate that our method maintains strong generalizability when trained on subsets generated by different models.}
  \label{tab:genimage-cross}
     % \vspace{-0.25 cm}
\end{table}
%-------------------- END OF TAB --------------------

\subsection{Fully Utilizing Different Layer Features}

Motivated by the mixture-of-experts paradigm~\cite{jacobs1991adaptive, jordan1994hierarchical} and our earlier observation that different Vision Transformer layers capture complementary information, we propose \texttt{MoLD}: \texttt{M}ixture-\texttt{o}f-\texttt{L}ayers for AI-generated image \texttt{D}etection. Our approach aims to effectively leverage the diverse features extracted across different layers through an adaptive gating mechanism, inspired by the concept of dynamic feature weighting introduced by \citet{jacobs1991adaptive}.

Given an input image $x$, we first extract the corresponding \cls token embedding from each ViT layer:
\begin{equation}
    \setlength\abovedisplayskip{8pt}%
    \setlength\belowdisplayskip{8pt}%
    F(x) = \{f^{(1)}(x), f^{(2)}(x), \dots, f^{(L)}(x)\},
\end{equation}
where $f^{(i)}(x) = (f_i \circ \cdots \circ f_1)(x)$ denotes the feature extracted from the $i$-th layer.

Next, each feature $f^{(i)}(x)$ is projected through a dedicated small neural network $g_i(\cdot)$ (implemented as a linear projection followed by GELU activation) into a common representation space:
\begin{equation}
    \setlength\abovedisplayskip{8pt}%
    \setlength\belowdisplayskip{8pt}%
    h_i = g_i(f^{(i)}(x)), \quad \textnormal{for } i = 1, \dots, L.
\end{equation}

The gating mechanism, implemented as a lightweight multi-layer perceptron $\mathcal{G}(\cdot)$, computes data-dependent layer weights $w_i$, dynamically determining the importance of each layer representation based on the top-layer embedding:
\begin{equation}
    \setlength\abovedisplayskip{8pt}%
    \setlength\belowdisplayskip{8pt}%
    \mathbf{w} = \operatorname{softmax}(\mathcal{G}(f^{(L)}(x))), \quad \mathbf{w} \in \mathbb{R}^{L}, \quad \sum_{i=1}^{L} w_i = 1.
\end{equation}

We then aggregate these layer representations using the learned weights into a single fused representation:
\begin{equation}
    \setlength\abovedisplayskip{8pt}%
    \setlength\belowdisplayskip{8pt}%
    h_{\text{fused}} = \sum_{i=1}^{L} w_i h_i.
\end{equation}

The fused representation $h_{\text{fused}}$ is subsequently passed through a linear classification head. We first compute a logit
\begin{equation}
    \setlength\abovedisplayskip{8pt}%
    \setlength\belowdisplayskip{8pt}%
    z(x) = W^\top h_{\text{fused}} + b,
\end{equation}
and obtain the predicted probability via a sigmoid activation:
\begin{equation}
    \setlength\abovedisplayskip{8pt}%
    \setlength\belowdisplayskip{8pt}%
    \hat{y}(x) = \sigma(z(x)).
\end{equation}

The feature projection networks $\{g_i\}_{i=1}^L$, the gating network $\mathcal{G}(\cdot)$, and the classification head (parameters $W, b$) are jointly trained using the binary cross-entropy loss:
\begin{equation}
    \setlength\abovedisplayskip{8pt}%
    \setlength\belowdisplayskip{8pt}%
    \mathcal{L} = -\frac{1}{N}\sum_{j=1}^{N} \left[ y_j \log \hat{y}(x_j) + (1-y_j) \log \big(1 - \hat{y}(x_j)\big) \right],
\end{equation}
where $N$ is the batch size and $y_j \in \{0, 1\}$ is the ground truth label.

This adaptive gating formulation allows us to dynamically capitalize on the distinct strengths of each transformer layer, significantly enhancing detection performance compared to methods relying solely on fixed or final-layer features. Despite its conceptual simplicity, our adaptive fusion strategy demonstrates substantial improvements in detecting AI-generated content. In subsequent sections, we evaluate \texttt{MoLD} across various pre-trained ViTs, including CLIP, highlighting its effectiveness and interpretability benefits.
\section{Experiments}
\label{sec:exp}

\subsection{Experimental Settings}

\textbf{Datasets.}
In this task of AI-generated image detection, both the training and test datasets are divided into two classes: real and fake. A binary classifier is then trained to function as the fake detector.
We utilized the ForenSynths~\cite{wang2020cnn} and GenImage~\cite{zhu2023genimage} datasets for the training and evaluation of our model. 
Since ForenSynths is widely used in AI-generated image detection research, we followed prior studies~\cite{jeong2022bihpf, jeong2022frepgan, tan2023learning, tan2024rethinking} by training our model on AI-generated images from four ProGAN classes (car, cat, chair, and horse), alongside real images from the LSUN dataset~\cite{yu2015lsun}. 
For the evaluation, we employed the ForenSynths test set and assessed our model on images generated by eight different generative models, primarily GAN-based, in line with previous works~\cite{tan2023learning, tan2024rethinking}. 
To specifically assess the performance on generated images from diffusion models, we incorporated the GenImage dataset, which has gained recent attention due to its inclusion of a diverse range of recent generative models.
Specifically, we trained our model on images generated by ADM, the type of first generative model in alphabetical order, using real images from ImageNet~\cite{deng2009imagenet}. 
We then tested the model on images from eight advanced generative models, predominantly diffusion-based, within the GenImage dataset. 
Since GenImage dataset includes both GAN and diffusion models, it is well-suited for testing a wide range of generative models. Therefore, unless otherwise specified, most experiments were conducted using the GenImage setup. 
Additional results on other training set of GenImage are provided Table~\ref{tab:genimage-cross} and Supp. B.

%-------------------- FIG: ROBUSTNESS --------------------
\begin{figure*}[ht]
    \centering
    \begin{minipage}[t]{0.71\textwidth}
        \centering
        \includegraphics[width=\textwidth]{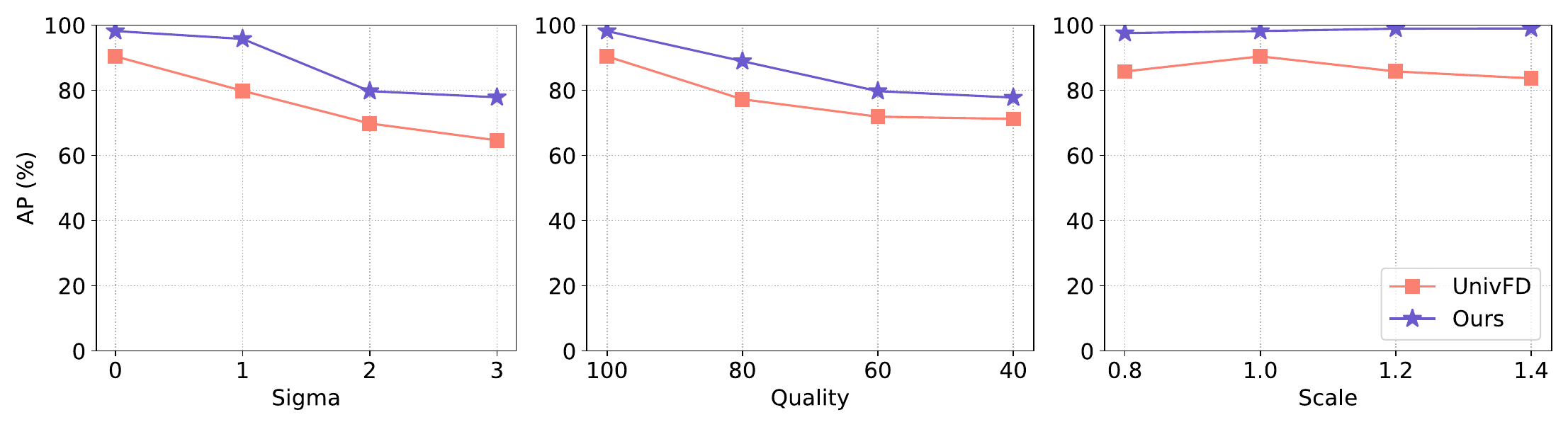}
        \subcaption{Perturbations}
        \label{fig:perturbations}
    \end{minipage}
    \hfill
    \begin{minipage}[t]{0.28\textwidth}
        \centering
        \includegraphics[width=\textwidth]{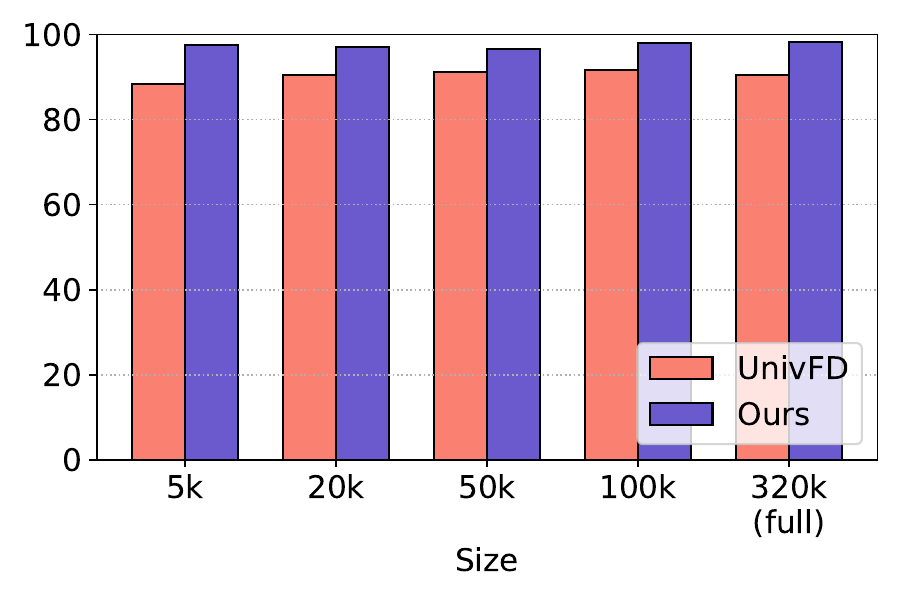}
        \subcaption{Train-data size}
    \end{minipage}
    \caption{\textbf{Robustness evaluation.} To assess the robustness of our approach in real-world scenarios, we conduct evaluations under various perturbations and training dataset sizes. Specifically, we apply perturbations including blur, JPEG compression, and scaling at different intensity levels (each sigma, quality and scale, respectively). Additionally, we vary the training dataset size from 5k to 320k, where 320k is the full dataset size of GenImage-ADM. Note that the total size includes both real and synthetic images. Our results demonstrate that the proposed method maintains robustness across a wide range of perturbations and dataset sizes.}
    \label{fig:robustness}
\end{figure*}
%-------------------- END OF FIG --------------------

\noindent\textbf{Baselines and metrics.}
For baselines, we compared our method against seven recent techniques: CNNDetection~\cite{wang2020cnn}, PatchFor~\cite{chai2020makes}, LGrad~\cite{tan2023learning}, UnivFD~\cite{ojha2023towards}, DIRE~\cite{wang2023dire}, NPR~\cite{tan2024rethinking}, and DRCT~\cite{chen2024drct}, which provides public training codebase. 
For DRCT, as they propose using two types of pre-trained networks: ConvNext-Base~\cite{liu2022convnet} and CLIP:ViT-L/14, we report results of both backbones.
All baseline models were trained using the settings specified in their respective original papers. 
As many prior studies in this field~\cite{wang2020cnn, chai2020makes, ojha2023towards} primarily focus on reporting average precision as a key evaluation metric, we also report average precision (\%) for most of our experiments. 
Additionally, for the main table, we include accuracy (\%) as another important evaluation metric.

\noindent\textbf{Implementation details.}
In order to show the effectiveness of our layer aggregation strategy, we adopt the majority of their experimental setup from \citet{ojha2023towards} and mainly utilize the CLIP:ViT-L/14 backbone. 
Specifically, we trained the model with an input resolution of 224 × 224, employing early stopping with patience of 5 epochs with learning rate decay.
All the hyperparameters, including the learning rate, were directly inherited from the UnivFD configuration.

\subsection{Main Results}

As demonstrated in Tables~\ref{tab:progan} and \ref{tab:genimage-adm}, our simple yet effective approach achieves significantly improved performance compared to the latest baselines.
Specifically, we achieve the highest average precision in both datasets: 99.5\% in ForenSynths and 98.2\% in GenImage. We also record the second best in ForenSynths and the highest average detection accuracy in GenImage.
While our method shares some methodological similarities with UnivFD, it fundamentally differs in its comprehensive utilization of features across multiple layers. 
This key distinction enables our approach to achieve significantly superior performance, demonstrating clear advantage over UnivFD. 

In particular, as shown in Table~\ref{tab:genimage-adm}, while many baseline methods struggle to achieve high detection performance, especially on Midjourney and Wukong, our method demonstrates consistently high detection performance. Furthermore, several baseline methods faced challenges in achieving satisfactory performance on SDv1.4 and SDv1.5, which can be attributed to the inherent differences between ADM, a pixel-space generative model trained on the dataset, and Stable Diffusion, which operates in the latent space. Despite these challenges, our method exhibited robust generalization across both types of generative models. 

This observation supports the claim made by UnivFD, which emphasizes that, rather than learning model-specific features, leveraging pre-trained ViT features enables better generalization across diverse generative models. Our results further substantiate this by showing that fully utilizing the rich feature representations learned from pre-trained ViTs allows our approach to outperform UnivFD, achieving superior performance in comparison.

Furthermore, GenImage offers training datasets derived from eight distinct generative models, enabling us to perform experiments using various source models for training. 
Therefore, we present the results of training on various training sets for each generative model in GenImage. The results are provided in Table~\ref{tab:genimage-cross}.
The results consistently demonstrate that our proposed \texttt{MoLD} exhibits robust generalization across different generative models, irrespective of the source image set utilized for training.

\subsection{Robustness}
% perturbation과 train dataset size에 대한 Robustness를 확인하기 위해 몇가지 실험을 진행하였다. ADM에서 학습을 하고 8개의 generative model에서 test한 후 mean AP를 report하였고, 가장 가까운 Baseline인 UnivFD와 비교하였다.
To assess robustness concerning perturbations (e.g., blur, jpeg compression, scale) and training dataset size (from 5k to full 320k), we conducted a series of experiments. The model was trained using ADM and tested on eight generative models, reporting the mean Average Precision (mAP). The results were then compared with UnivFD, which is methodologically the closest baseline and performing the second-best next to ours in Table~\ref{tab:genimage-adm}.

\noindent\textbf{Robustness to perturbations.}
In the context of fake image detection, adversaries can evade detection by applying perturbations to images. 
Additionally, given that JPEG compression is commonly applied on the web, we evaluated the robustness of our method against such perturbations. 
Furthermore, we examined whether detection performance varies with scale due to resizing effects.  
For these evaluations, we conducted experiments under various conditions. 
To assess the impact of blur, we applied Gaussian blur with different sigma values. 
For JPEG compression, we varied the compression strength by adjusting the JPEG quality factor. 
Lastly, for scale variations, we resized images to different scales using the default bilinear interpolation function in the torchvision resize module. 

The results of these experiments are presented in Figure~\ref{fig:perturbations}.  
While detection performance decreases as perturbation intensity increases in both Gaussian blur and JPEG compression, our method consistently achieves high performance across different levels of blur and compression. 
Regarding scale variations, both UnivFD and our approach demonstrated a certain level of robustness. However, while UnivFD exhibited a slight performance drop, our method remained nearly unaffected.  
In summary, our approach demonstrates robustness against various levels of perturbations, including blur, JPEG compression, and scale variations.

%------------------------------------------
\begin{figure}[t]
    % \centering
    \includegraphics[width=\linewidth]{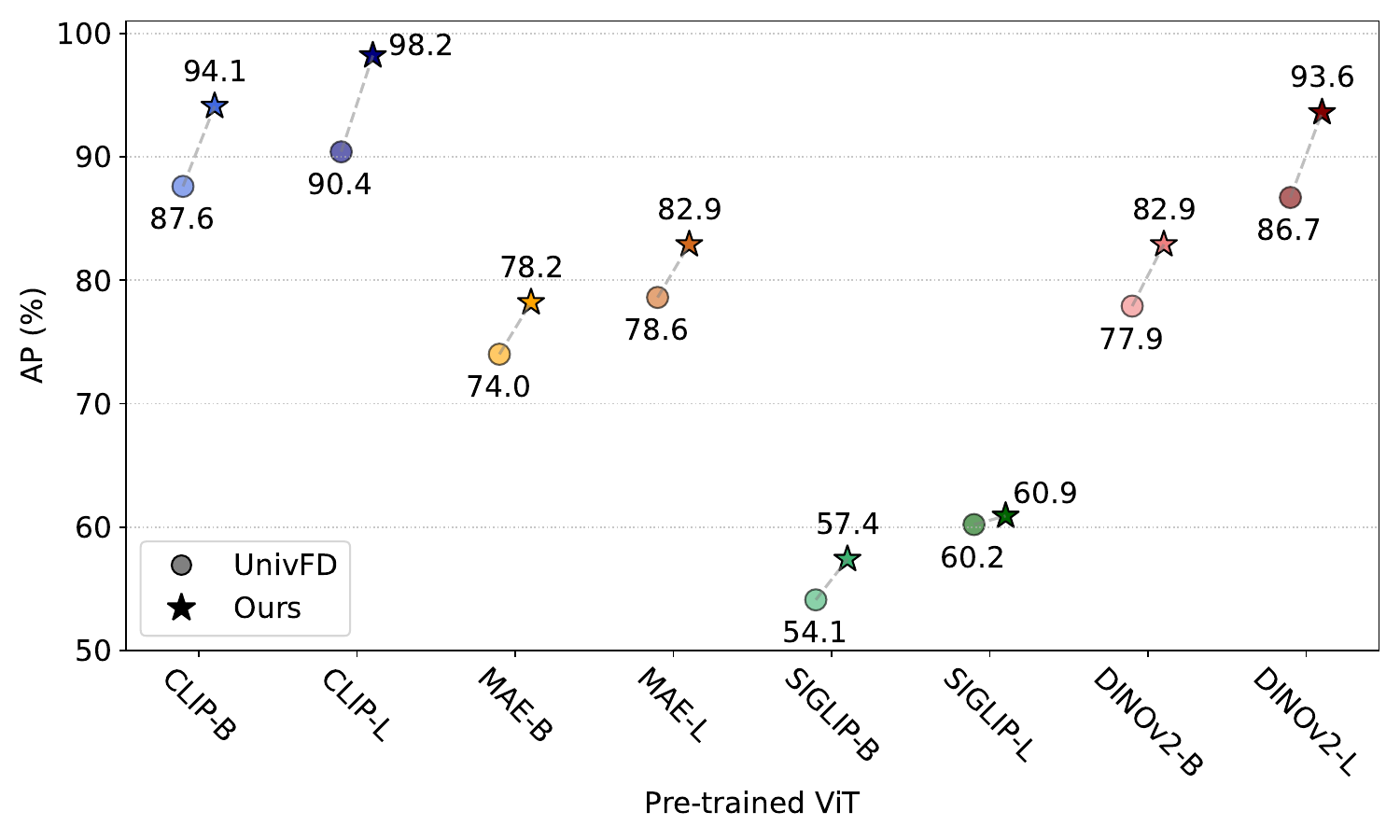}
    \vspace{-1mm}
    \caption{\textbf{Performance across different pre-trained Vision Transformers (ViTs).} Our approach can be seamlessly integrated into various pre-trained ViTs, including CLIP \cite{rombach2022high}, MAE (Masked Autoencoders) \cite{he2022masked}, SigLIP \cite{zhai2023sigmoid}, and DINOv2 \cite{oquab2023dinov2}. We observe consistent performance improvements across different backbones, demonstrating the effectiveness of fully leveraging prior features for AI-generated image detection. Additionally, larger models tend to yield superior performance, with CLIP ViT-L/14 achieving the best results, followed by DINOv2-Large.}
    \label{fig:backbone}
\end{figure}
%------------------------------------------

\noindent\textbf{Robustness to train-data size.}
To evaluate the robustness with respect to training dataset size, we conducted experiments using randomly selected subsets of the GenImage ADM dataset (full 320k). Training was performed on progressively larger subsets, starting from a minimal size of 5k, where the dataset size refers to the total number of real and fake images combined. 

The results, presented in Figure~\ref{fig:robustness} (b), demonstrate that even with a very small dataset of 5k samples, the model achieves remarkably high performance. Overall, the findings indicate that the model exhibits robustness with minimal sensitivity to dataset size.

\subsection{Expanding to Different Pre-trained ViTs}
Throughout this paper, the experiments were conducted using the same backbone as UnivFD, which leverages the features of CLIP:ViT-L/14. 
However, our method is not restricted to CLIP ViT; it can be applied to various pre-trained Vision Transformers, fully leveraging prior features for AI-generated image detection.
Therefore, we evaluated its applicability to not only CLIP but also Masked AutoEncoder (MAE)~\cite{he2022masked}, SigLIP~\cite{zhai2023sigmoid}, and DINOv2~\cite{oquab2023dinov2}, that encompass different pre-train data and training techniques. 
For comparison, we evaluated UnivFD under the same conditions, replacing only the backbone while maintaining the use of the last-layer features. 
Similarly, our \texttt{MoLD} was applied with different backbones, utilizing features from all layers. For SigLIP, which does not have a \cls token, we used pooled features instead.  
The results of these experiments are presented in Figure~\ref{fig:backbone}, demonstrating that our approach is applicable to a wide range of pre-trained ViTs.
%------------------------------------------
\begin{figure}[t]
    \centering
    \includegraphics[width=0.85\linewidth]{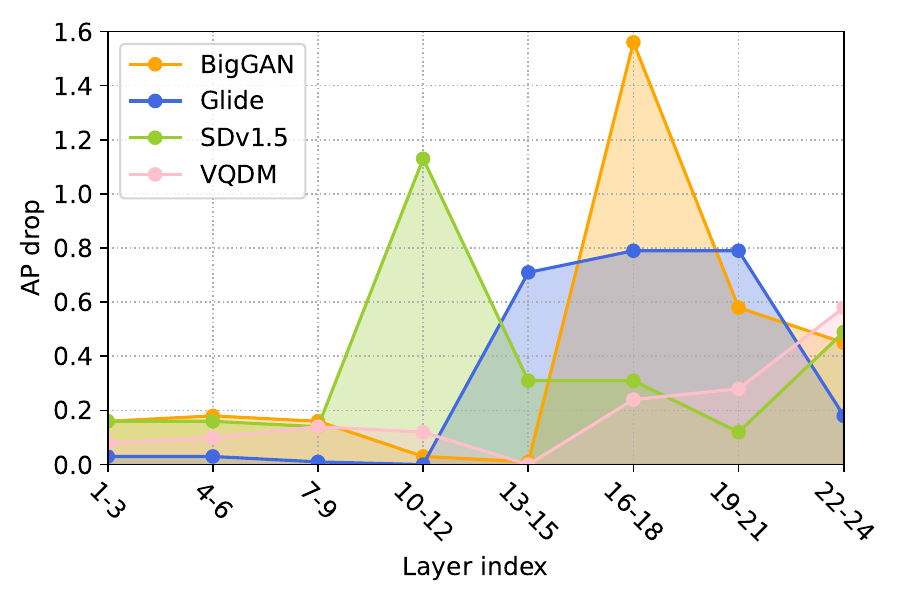}
    \vspace{-2mm}
    \caption{\textbf{Ablation on layer contributions.} To assess the impact of each layer's features on the performance of the fake image detector, we conducted an ablation study by excluding the features from specific layers of the trained detector. The results demonstrate that all layers contribute to the model's performance, with the degree of contribution varying across different settings. This highlights the importance of considering the full set of layer features, thereby supporting the rationale for our approach.}
    \label{fig:layer}
\end{figure}
%------------------------------------------

These experiments yielded several key observations.  
(1) Overall, UnivFD exhibited a certain level of detection performance across nearly all pre-trained ViTs, indicating that many pre-trained ViTs provide useful features for fake image detection.  
(2) Compared to UnivFD, our method consistently outperformed across diverse pre-trained ViTs. This suggests that our \texttt{MoLD} enables a more comprehensive utilization of the features provided by pre-trained ViTs, demonstrating its effectiveness for the AI-generated image detection task.  
(3) Model size of pre-trained ViT affects performance: in all ViT settings in the figure, larger models consistently outperformed their base counterparts.
(4) There are distinct performance differences among ViT variants, likely due to differences in their pre-trained datasets, architectural details, and training methodologies. Notably, CLIP:ViT-L, which was used in the UnivFD setting, achieved the highest performance, followed by DINOv2-Large. However, SigLIP exhibited relatively poor performance, which may be attributed to its use of pooled features instead of a \cls token.

In summary, while various pre-trained Vision Transformers (ViTs) with different architectures and sizes provide valuable features for fake image detection, our approach effectively utilizes these features across all layers, resulting in a performance improvement and demonstrating its efficacy in detecting AI-generated images.

\subsection{Analyzing the Results}
\noindent \textbf{Which layer contributes the most?}
We conducted an ablation study by removing features from specific layers and measuring the performance drop. 
Initially, we removed one layer at a time, but with 23 layers remaining, the differences were small. 
To get clearer results, we removed features from three layers at once. 
As shown in Figure~\ref{fig:layer}, the performance drop varied across test subsets: layers 16-18 had the most significant drop for the BigGAN dataset, while layers 10-12 had the second-largest drop for the SDv1.5 dataset. 
To summarize, this confirms the distinct and varying roles of different layers in detection performances. It aligns with our earlier observation on the importance of integrating features from all layers.

%------------------------------------------
\begin{figure}[t]
    \centering
    \includegraphics[width=0.95\linewidth]{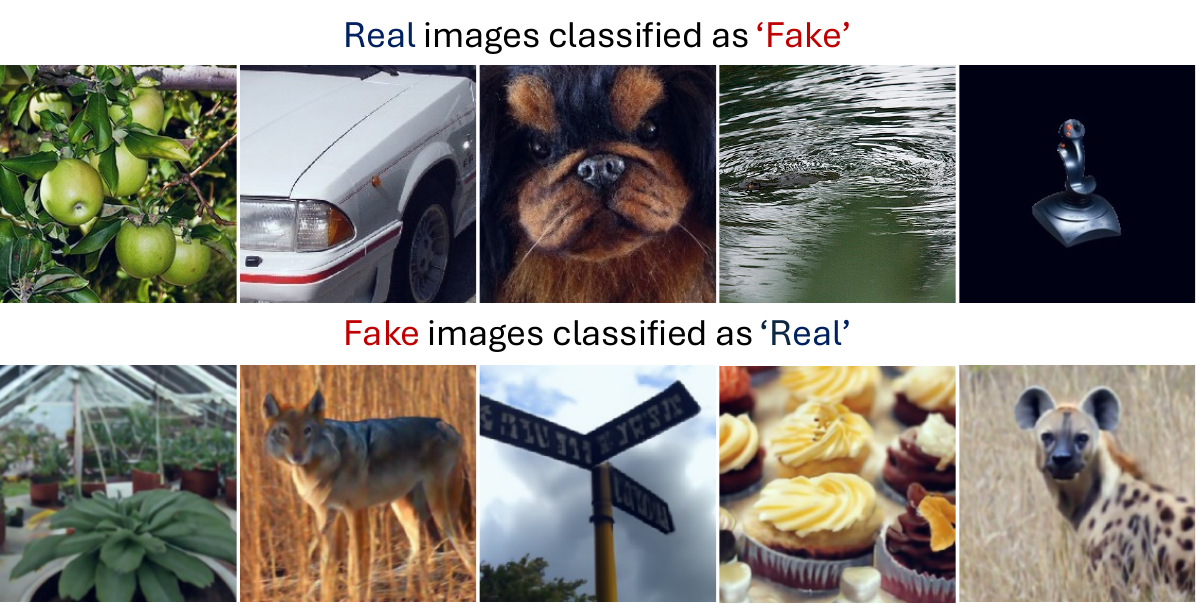}
    \caption{\textbf{Misclassified examples.} We show instances of misclassified images with the lowest confidences from our AI-generated image detection model. The examples include both false negatives (authentic images incorrectly classified as fake) and false positives (AI-generated images incorrectly classified as authentic). These samples demonstrate no apparent contextual anomalies upon visual inspection.} \vspace{-0.2mm}
    \label{fig:wrong}
\end{figure}
%------------------------------------------

\noindent \textbf{Examples of hard samples.}
We conducted an analysis of the types of images that presented the greatest challenges for our fake image detector, specifically focusing on those where the classifier demonstrated the lowest confidence scores. 
This involved examining real images that were misclassified as fake (false positives) and fake images that were misclassified as real (false negatives). 
Examples of these misclassified samples are shown in Figure~\ref{fig:wrong}.

Upon visual inspection, no clear patterns were identifiable among these misclassified samples, suggesting that the classifier's performance has reached a level that is beyond reliable detection by human observation. 
However, we did observe that some real images from the GenImage dataset, sourced from ImageNet and stored in JPEG format, exhibited noticeable compression artifacts. Additionally, certain images contained wave-like distortions, while others appeared to have been extracted from their original backgrounds.

\section{Discussion}
\label{sec:discussion}

\noindent\textbf{Limitations \& Future directions.}
Our research builds on recent work using CLIP-ViT and focuses on improving the usability of pre-trained Vision Transformers. Although our approach can also be applied to CNN-based detectors, we focus on ViTs due to UnivFD's reports that ViT-based CLIP outperforms ResNet-based CLIP for detecting AI-generated images.
Even though our approach leads to significant performance improvements, certain samples—especially from specific generative models—remain difficult to detect. For example, the ForenSynths-DeepFake and GenImage-Midjourney subset show lower detection rates across most of  baseline approaches, including ours. Solving this issue may require strategies specific to these models.
It’s also worth noting that our approach, along with all the baseline methods, is designed to detect fully-generated images from generative models. Detecting AI-edited images will require further research and different techniques. We believe that more explainable research is needed to identify the unique features that distinguish real from synthetic images, which will help improve detection performance beyond current benchmarks.

\noindent\textbf{Societal impact.}
The rapid advancement of generative models has given rise to significant concerns regarding their potential negative impacts. 
Notably, privacy infringements resulting from fabricated images and the dissemination of false information are critical issues that must be addressed. This research investigates how content generated by such models can be analyzed using features from pre-trained models, offering a promising approach for a wide range of forgery detection methods. 
To ensure secure and trustworthy AI systems, it is essential to establish robust safeguards against these emerging threats.
\section{Conclusion}
\label{sec:conc}

In this paper, we addressed the research question of whether the commonly adopted approach—utilizing only the last-layer representation—is truly optimal for AI-generated image detection. 
Through a comprehensive layer-wise analysis, we demonstrated that earlier-layer features contribute valuable information and that each layer offers distinct decision-making insights for this task. 
Based on these findings, we proposed a simple yet effective layer aggregation strategy, which achieved strong performance across diverse scenarios. 
Our research offers fresh perspectives on forgery detection, highlighting the scalability of our approach and underscoring the importance of fully leveraging the rich feature representations inherent in pre-trained Vision Transformers.
We hope that our findings serve as a valuable contribution to the field and inspire further investigations into this promising direction.
{
    \small
    \bibliographystyle{main}
    \bibliography{main}
}
\clearpage
\renewcommand{\thesubsection}{\Alph{subsection}}

\section*{Supplementary Material}

%%%%%%%%% %%%%%%%%% %%%%%%%%% %%%%%%%%% %%%%%%%%% %%%%%%%%% %%%%%%%%% 
\subsection{More Results on Prior Analysis}
\subsubsection{Impact of high-level semantics on each layer}
\label{supp:a3}
%------------------------------------------
\begin{figure}[h!]
    \centering
    \includegraphics[width=0.9\linewidth]{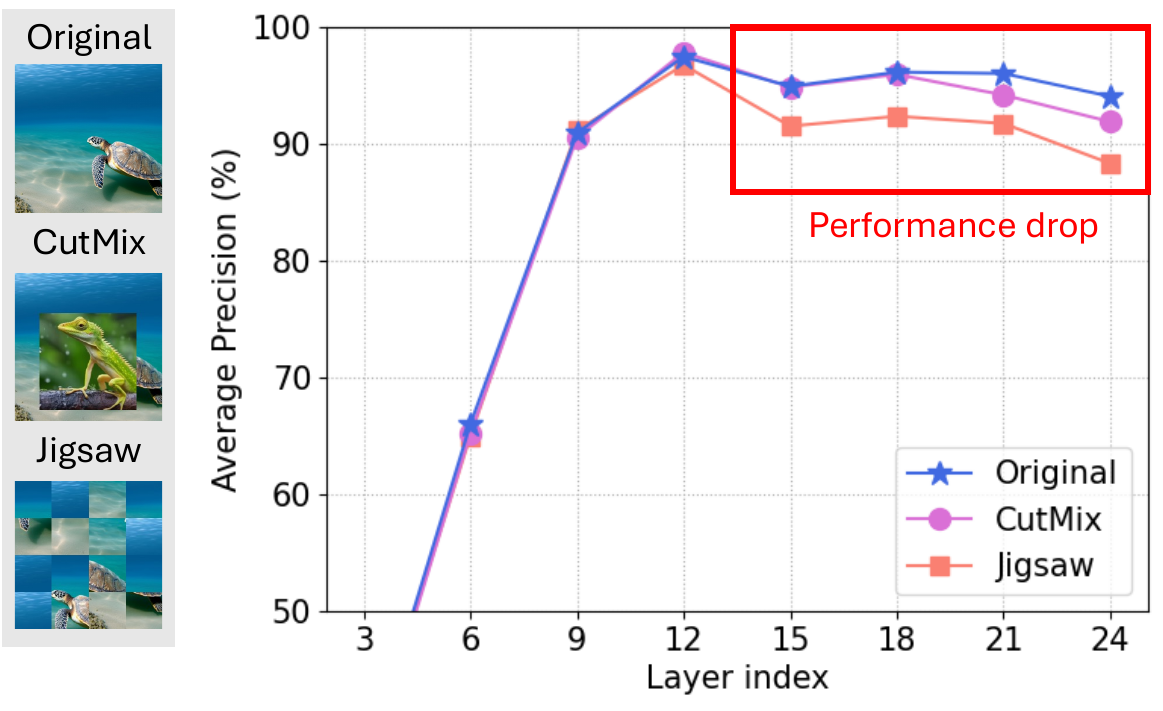}
    \caption{\textbf{Impact of semantic transformation on each layer.} We analyze how semantic modifications affect fake image detection by applying CutMix and Jigsaw transformations. In CutMix, real images are mixed with real images and fake images with fake images, keeping the true labels unchanged. To understand how different network layers respond to high-level semantics, we train a fake image detector using features from individual layers and report their Average Precision. The results show that higher layers are more affected by semantic changes, leading to lower detection performance. Meanwhile, mid-layer features achieve better detection accuracy than both early and high-level features, making them more effective for distinguishing fake images.}
    \label{fig:supp_semantic}
\end{figure}
%------------------------------------------
In the experiment conducted to investigate the relationship between layer-specific features and high-level semantics, we aimed to test the degree to which each layer’s features are influenced by semantic information. To achieve this, we modified the semantics of the original images and assessed the impact on the model's performance. Specifically, we applied CutMix \cite{yun2019cutmix} and Jigsaw transformations, as illustrated in Figure~\ref{fig:supp_semantic}. Since fake image detection should ideally be invariant to changes in semantics, the model should still classify images as fake, even when their semantics are altered through transformations such as shuffling (Jigsaw) or mixing (CutMix).

The experimental results demonstrate that, while most layers maintain consistent performance, later layers exhibit a noticeable decline in accuracy. This suggests that the layers closer to the final output are more sensitive to semantic information, while earlier layers are less influenced by it. This finding aligns with established observations in the transfer learning literature \cite{yosinski2014transferable}, which indicate that earlier layers of a pre-trained network capture more general features, whereas later layers become increasingly specialized to the specific task of pre-training. Based on this framework, it is reasonable to infer that the later layers retain semantic information learned during CLIP training. Consequently, for the task of fake image detection, it may be more advantageous to utilize features from the earlier layers, which are less sensitive to high-level semantics.

\subsubsection{Performance of layer-wise classifiers}
\label{supp:a1}
%-------------------- FIG --------------------
\begin{figure}[ht!]
    \centering
    \begin{minipage}[t]{0.9\linewidth}
        \centering
        \includegraphics[width=\textwidth]{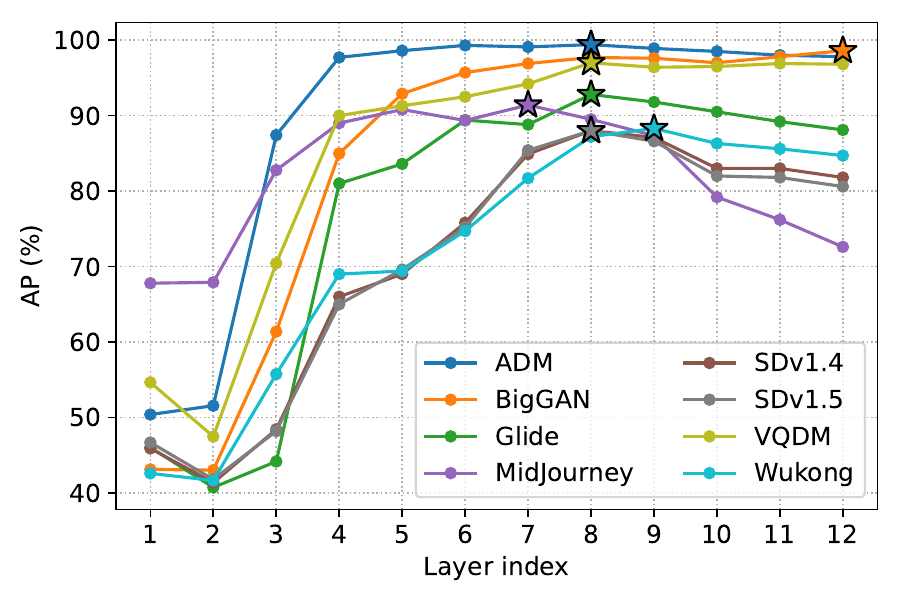}
        \subcaption{GenImage (CLIP:ViT-B/16)}
    \end{minipage}
    \vfill
    \begin{minipage}[t]{0.9\linewidth}
        \centering
        \includegraphics[width=\textwidth]{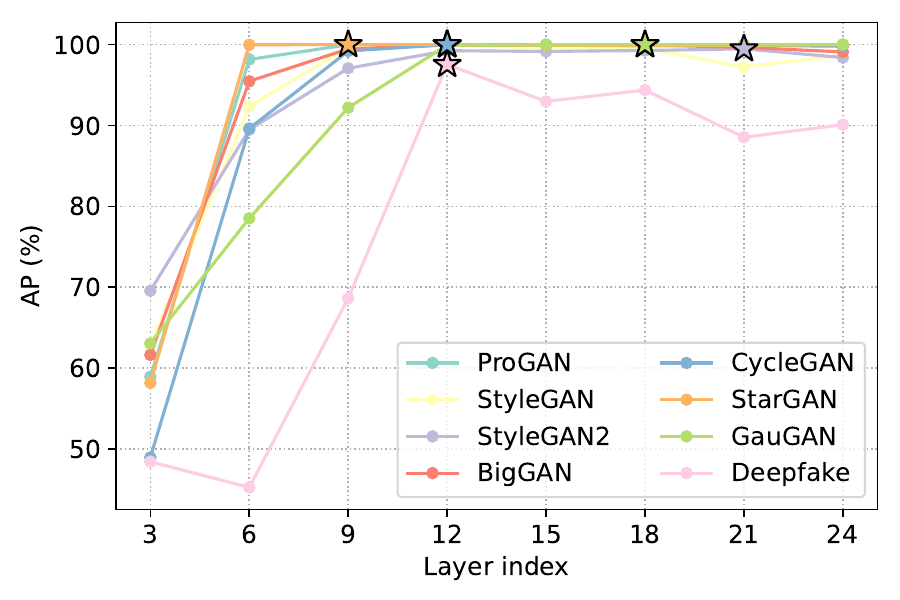}
        \subcaption{ForenSynths (CLIP:ViT-L/14)}
    \end{minipage}
    \caption{\textbf{Layer-wise results on different settings.} We plot the detection performance of layer-wise classifiers trained on diverse settings, in average precision. Results indicate that the earlier layers compared to final layers generally achieves better detection paerformance.}
    \label{fig:supp_layerwise}
\end{figure}

%-------------------- END OF FIG --------------------
In the main paper, we primarily investigated CLIP:ViT-L/14, which was the backbone used by \citet{ojha2023towards}. Given that CLIP:ViT-L/14 is a large model with 24 transformer layers, we grouped every three consecutive layers into one. To assess whether the findings from the main paper are broadly applicable, we replicated the experiment using every layer of CLIP:ViT-B/16 and a different dataset. The results are presented in Figure \ref{fig:supp_layerwise}. 

Consistent with the findings in the main paper using CLIP:ViT-L/14, we observed that the highest detection performance is typically achieved in the layers preceding the final layer. This trend also holds for the ForenSynths dataset, where features from the early to mid-layers contribute to higher detection rates compared to those from the final layer.

\subsubsection{Investigating layer-wise decisions}
\label{supp:a2}
In the main paper, we present results from the CLIP:ViT-L/14 model. Here, we replicate the experiment to show the layer-wise overlap for CLIP:ViT-B/16. The results in the main paper are derived from the GenImage-Glide dataset, whereas in this supplementary section, we provide results based on a different dataset, GenImage-Midjourney.

The findings reveal that the overlap between the classifiers at each layer varies across layers. By examining each layer individually, we observe more overlap than in the main figure. However, despite this increased overlap, it remains relatively low overall, suggesting that each layer's classifier makes distinct decisions.

In conclusion, our analysis demonstrates that the features extracted from each layer contribute differently to the overall model's output. This variability underscores the importance of our Layer-wise Attention Distillation (LAD) method.
%------------------------------------------
\begin{figure}[t!]
    \centering
    \includegraphics[width=0.8\linewidth]{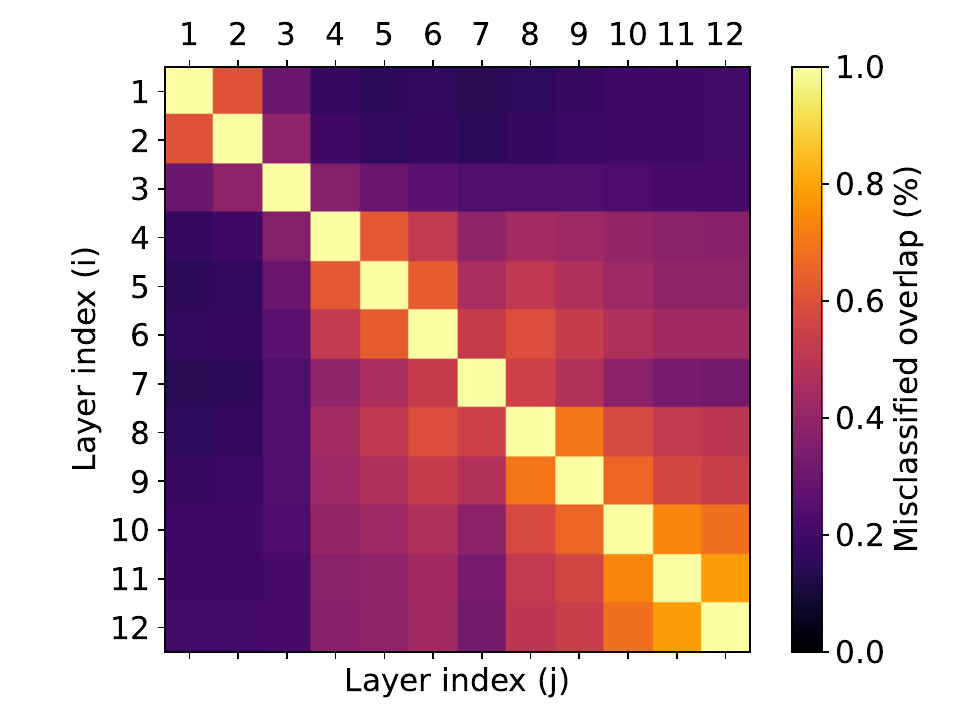}
    \caption{\textbf{Overlap of misclassified examples in CLIP-ViT-Base.} To analyze the decision-making process of classifiers trained on features from different layers, we measure the proportion of commonly misclassified samples. The results indicate that the overlap in misclassified samples across layers is relatively low, suggesting that each layer captures distinct aspects of the input data.}
    \label{fig:supp_wrong}
\end{figure}
%------------------------------------------

\subsection{More Results}
\label{supp:b}
%--------------------TAB: GENIMAGE-BIGGAN --------------------
\begin{table*}[!ht]
    \centering
% \resizebox{0.8\linewidth}{!}{
    \begin{tabular}{l r r r r r r r r | r}
\toprule
& \ct{ADM} & \ct{BigGAN} & \ct{GLIDE} & \ct{Midjourney} & \ct{SDv1.4} & \ct{SDv1.5} & \ct{VQDM} & \ct{Wukong} & \multicolumn{1}{|c}{Mean} \\ \midrule
            CNNDetection & 61.9 & 100.0 & 93.8 & 54.1 & 47.8 & 47.7 & 64.9 & 40.6 & 63.1 \\
            PatchFor & 81.9 & 100.0 & 98.0 & 78.8 & 55.2 & 55.8 & 66.2 & 59.2 & 74.6 \\
            Lgrad & 61.0 & 100.0 & 91.3 & 62.2 & 48.8 & 49.0 & 58.0 & 43.4 & 64.5 \\
            UnivFD & 82.5 & 99.9 & 54.3 & 75.7 & 82.3 & 82.3 & 96.5 & 83.5 & 82.1 \\
            DIRE & 59.5 & 100.0 & 95.0 & 60.0 & 49.1 & 49.3 & 54.3 & 45.5 & 64.6 \\
            NPR & 59.8 & 100.0 & 98.1 & 61.6 & 55.5 & 55.9 & 64.3 & 46.0 & 67.7 \\
            DRCT (Conv-B) & 55.7 & 100.0 & 67.6 & 56.5 & 49.9 & 50.0 & 58.6 & 42.0 & 60.0 \\
            DRCT (CLIP-L) & 90.6 & 100.0 & 94.9 & 65.7 & 72.6 & 71.5 & 92.5 & 67.7 & 81.4 \\ \midrule
            \rowcolor{gray!25} LAD (Ours) & 78.4 & 99.9 & 95.5 & 76.1 & 94.0 & 94.5 & 97.5 & 90.7 & \textbf{90.1} \\
        \bottomrule
        \end{tabular}
% }
        \caption{\textbf{Evaluation on GenImage-BigGAN.} We present the detection performance of baseline methods and our approach in terms of average precision (AP). Each detector is trained on the GenImage-BigGAN training set, which follows GenImage-ADM in the alphabetical sequence of generative models. The results include average precision and mean average precision (mAP). Our LAD method achieves the highest detection performance, while UnivFD yields the second-best result.}
  \label{supp:tab-biggan}
\end{table*}
%-------------------- END OF TAB --------------------
We train our AI-generated image detector using the GenImage-ADM dataset, which is the first generative model in the GenImage collection. We compare the performance of our model with baseline approaches, as well as with models trained on the subsequent dataset in alphabetical order, GenImage-BigGAN (see Table \ref{supp:tab-biggan}). As shown in the table, our LAD model achieves the best performance, surpassing the second-best model by over 8\% in average precision. This improvement is more pronounced than the results obtained using the GenImage-ADM dataset, as reported in the main paper's table. We hypothesize that this enhanced performance is due to the greater differences between BigGAN-generated images and the diffusion model-generated images in the test set, which predominantly consists of diffusion model samples. This demonstrates the superior generalization capability of our approach.

\subsection{Additional Analyses}
\subsubsection{Attention distance of CLIP-ViT}
\label{supp:c1}
%------------------------------------------
\begin{figure}[t!]
    \centering
    \includegraphics[width=0.8\linewidth]{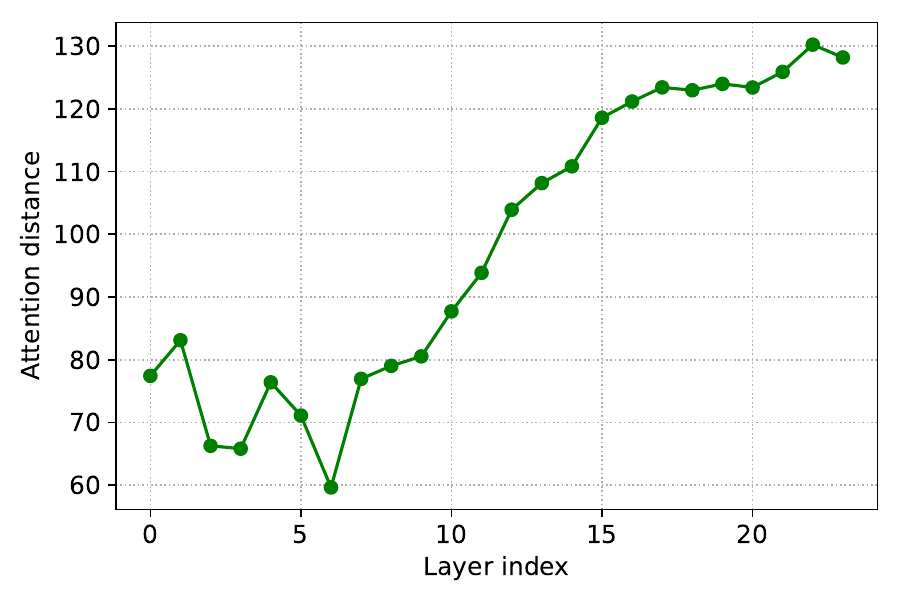}
    \caption{\textbf{Attention distance of CLIP:ViT-L/14.} The attention distances for 500 images from ForenSynths are measured and averaged for plotting. CLIP:ViT-L/14 is used as the backbone, resulting in 24 data points.}
    \label{fig:supp_attn}
\end{figure}
%------------------------------------------
To examine the local and global receptive fields of each transformer layer, we calculate the attention distance, as proposed in previous works \cite{raghu2021vision, park2023self}. Specifically, the attention distance is determined by the pixel distance, weighted by the corresponding attention scores. As illustrated in Figure \ref{fig:supp_attn}, the attention distance increases with the layer index. This observation indicates that lower layers focus primarily on local regions, mid-level layers capture both local and global information, and higher layers predominantly attend to global contexts.

\subsubsection{t-SNE plot of each layer}
\label{supp:c2}
%------------------------------------------
\begin{figure*}[ht!]
    \centering
    \includegraphics[width=0.975\linewidth]{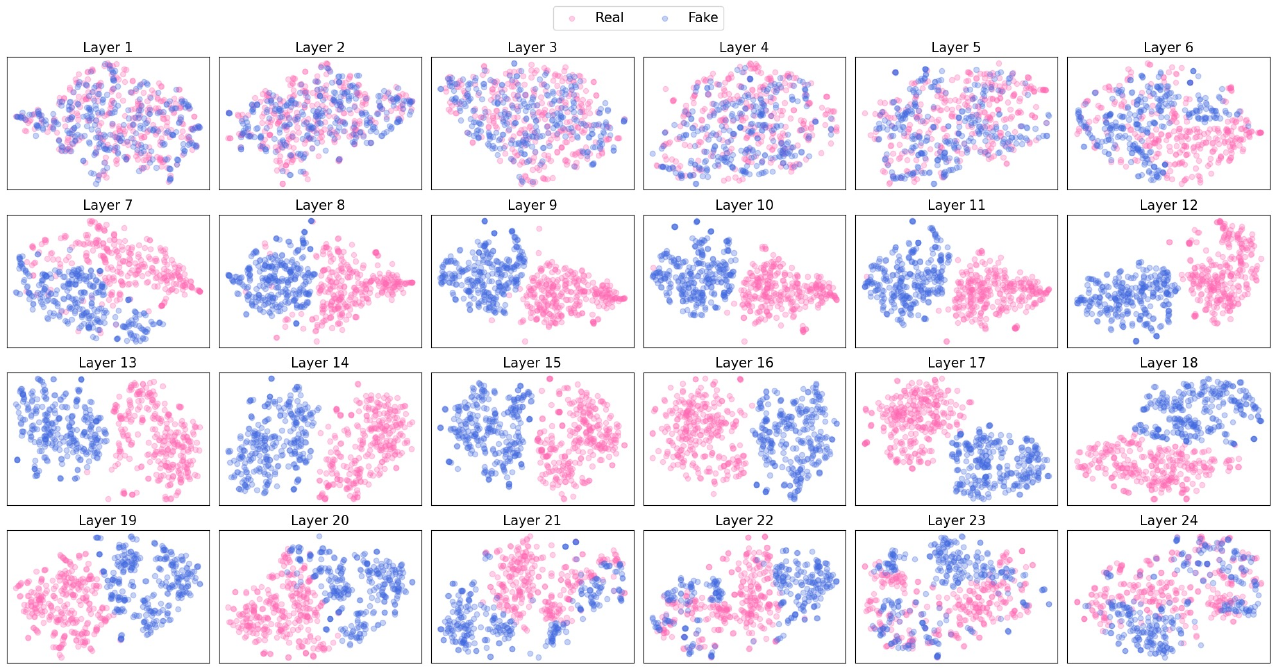}
    \caption{\textbf{Layer-wise t-SNE plot.} We present the t-SNE visualization of features extracted from each layer of the ForenSynths BigGAN model, using a test set of 600 images (300 real and 300 fake). The results suggest that the middle layers exhibit a stronger ability to distinguish between real and fake images compared to the early and final layers.}
    \label{fig:supp_tsne}
\end{figure*}
%------------------------------------------
We present the t-SNE \cite{van2008visualizing} plots of each layer of CLIP:ViT-L/14 in Figure \ref{fig:supp_tsne}. As illustrated in the figure, the real and fake images are more distinguishable in the early and middle layers compared to the layers at the extremes. Specifically, in the initial layers (1 to 5), the real and fake images are largely indistinguishable, whereas in the final layers (21 to 24), the features become more separable, though still not as distinct as in the middle layers. This observation is consistent with and supports the results of our layer-wise detection performance discussed in the previous analysis of the main paper.

\subsubsection{Effect of number of linear layers}
\label{supp:c2}
%--------------------TAB: GENIMAGE-BIGGAN --------------------
\begin{table*}[!ht]
    \centering
\resizebox{0.95\linewidth}{!}{
    \begin{tabular}{l  r r r r r r r r r r r r r r r r | r r}
    \toprule
       \multirow{2}*{Method} & \multicolumn{2}{c}{ADM}& \multicolumn{2}{c}{BigGAN}& \multicolumn{2}{c}{GLIDE}& \multicolumn{2}{c}{Midjourney}& \multicolumn{2}{c}{SDv1.4}& \multicolumn{2}{c}{SDv1.5}& \multicolumn{2}{c}{VQDM}& \multicolumn{2}{c|}{Wukong}& \multicolumn{2}{c}{Mean}\\

         \cline{2-19} ~   & \ct{ACC} & \ct{AP} & \ct{ACC} & \ct{AP} & \ct{ACC} & \ct{AP} & \ct{ACC} & \ct{AP} & \ct{ACC} & \ct{AP} & \ct{ACC} & \ct{AP} & \ct{ACC} & \ct{AP}  & \ct{ACC} & \ct{AP}  & \multicolumn{1}{|c}{ACC} & \ct{AP} \\ \midrule
        UnivFD & 90.6 & 97.1 & 91.7 & 99.1 & 79.0 & 88.2 & 61.8 & 69.5 & 80.3 & 89.2 & 79.5 & 88.4 & 90.8 & 98.5 & 84.8 & 93.3 & 82.3 & 90.4 \\
         UnivFD (2-layer MLP) & 94.8 & 99.3 & 95.4 & 99.6 & 87.1 & 94.8 & 59.3 & 69.8 & 75.9 & 87.6 & 74.4 & 86.6 & 92.9 & 98.4 & 80.5 & 91.3 & 82.6 & 90.9 \\
        \midrule
        \rowcolor{gray!25} Ours & 99.3 & 100.0 & 83.3 & 97.9 & 91.1 & 99.0 & 76.3 & 95.2 & 88.2 & 98.5 & 87.0 & 98.3 & 93.5 & 99.2 & 86.5 & 98.1 & \multicolumn{1}{|r}{\textbf{88.2}} & \textbf{98.2} \\
    \bottomrule
    \end{tabular}
}
        \caption{\textbf{Evaluation of UnivFD in 2-layer MLP.} We implement 2-layer classification head to UnivFD in order to examine the effectiveness of 2-layer MLP on its performance. The results are negligability different with a little performance increase. Still, our LAD noticeably outperforms UnivFD, demonstrating the effectiveness of layer aggregation strategy of our approach. Note that all the other values except the second row is same to GenImage-ADM table in the main paper.}
  \label{supp:tab-univfd-fc2}
\end{table*}
%-------------------- END OF TAB --------------------
Strictly speaking, UnivFD employs a single-layer MLP, whereas our approach utilizes a two-layer MLP with a GELU activation function \cite{hendrycks2016gaussian} between the layers. Since UnivFD relies solely on a single linear layer, all results reported in the main paper are based on its original implementation. This architectural difference raises the question of whether the observed performance gap is due to the additional non-linearity in our model. To investigate this, we re-implemented UnivFD using a two-layer MLP with GELU and present the results in Table \ref{supp:tab-univfd-fc2}. As shown in the table, the performance difference is negligible, indicating that the superior performance of our LAD approach is not attributable to the two-layer MLP architecture but rather to the effectiveness of our method in fully leveraging the features from different layers of pre-trained CLIP-ViT.

\subsubsection{Consistency analysis}
\label{supp:c3}
%------------------------------------------
\begin{figure}[ht!]
    \centering
    \includegraphics[width=0.8\linewidth]{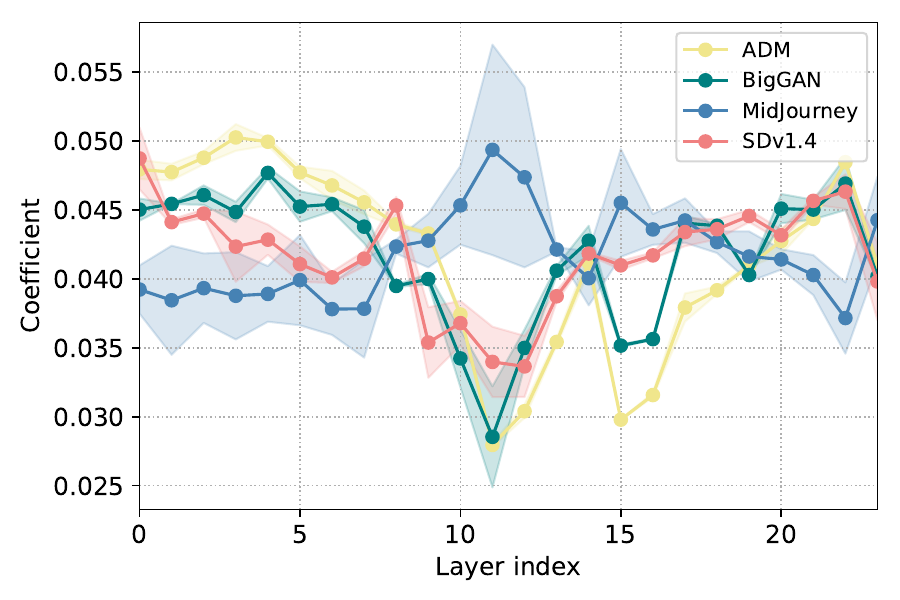}
        \caption{\textbf{Consistency in training.} Each model is trained three times, and we report the mean and standard deviation of the coefficients for each layer. The results demonstrate consistent layer-wise utilization across the different training sets for the detectors.}
    \label{fig:supp_cons}
\end{figure}
%------------------------------------------
We assess the consistency of our LAD model by training it three times with different random seeds and plotting the mean and standard deviation of the coefficients for each layer. It is important to note that this analysis does not directly reflect the importance of each layer, as it includes biases such as the mean or norm of the features for each layer. However, this evaluation demonstrates the consistency of our LAD model, illustrating that the network effectively learns to utilize all features from each layer, even when trained with different random initializations.

\end{document}